\documentclass[runningheads]{llncs}
\usepackage{svg}
\usepackage[T1]{fontenc}

\usepackage{graphicx}

\begin{document}

\title{Privacy-Preserving in Medical Image Analysis: A Review of Methods and Applications}

\titlerunning{Privacy-Preserving in Medical Image Analysis: A Review}

%\author{First Author\inst{1}\orcidID{0000-1111-2222-3333} \and
%Second Author\inst{2,3}\orcidID{1111-2222-3333-4444} \and
%Third Author\inst{3}\orcidID{2222--3333-4444-5555}}


%\institute{Princeton University, Princeton NJ 08544, USA \and
%Springer Heidelberg, Tiergartenstr. 17, 69121 Heidelberg, Germany

\author{Yanming Zhu \and Xuefei Yin \and Alan Wee-Chung Liew \and Hui Tian}
\institute{School of Information Communication and Technology, Griffith University, Gold Coast 4215, Australia %\\
\email{\{yanming.zhu,x.yin,a.liew,hui.tian\}@griffith.edu.au}}

\maketitle         

\begin{abstract}
%in 150--250 words.
With the rapid advancement of artificial intelligence and deep learning, medical image analysis has become a critical tool in modern healthcare, significantly improving diagnostic accuracy and efficiency. However, AI-based methods also raise serious privacy concerns, as medical images often contain highly sensitive patient information. This review offers a comprehensive overview of privacy-preserving techniques in medical image analysis, including encryption, differential privacy, homomorphic encryption, federated learning, and generative adversarial networks. We explore the application of these techniques across various medical image analysis tasks, such as diagnosis, pathology, and telemedicine. Notably, we organizes the review based on specific challenges and their corresponding solutions in different medical image analysis applications, so that technical applications are directly aligned with practical issues, addressing gaps in the current research landscape. Additionally, we discuss emerging trends, such as zero-knowledge proofs and secure multi-party computation, offering insights for future research. This review serves as a valuable resource for researchers and practitioners and can help advance privacy-preserving in medical image analysis.

\keywords{Privacy-preserving \and Medical image analysis \and Federated learning \and Differential privacy \and Encryption \and Image obfuscation/deformation.}
\end{abstract}

\section{Introduction}
With the rapid advancement of medical technology, medical image analysis has become vital in modern healthcare. Through images like computed tomography (CT), magnetic resonance imaging (MRI), ultrasound, and pathology, clinicians can diagnose diseases more accurately, assess treatment effectiveness, and even predict potential health risks \cite{hussain2022modern}. Artificial intelligence (AI) and deep learning (DL) have shown immense potential in medical image analysis, driving advancements in tumor detection, organ segmentation, pathology, and more \cite{pinto2023artificial}. These advancements not only accelerate data processing but also reduce healthcare professionals' workloads, addressing the shortage of specialized medical staff. Additionally, with the rise of the Internet of Things (IoT) and telemedicine, the collection and analysis of medical images have expanded.

However, the widespread use of AI in medical image analysis raises significant data privacy concerns \cite{khalid2023privacy}. Medical images contain sensitive personal information, including physiological characteristics, medical histories, and diagnostic results. Data breaches during transmission, storage, or analysis can lead to serious privacy violations or identity theft. As medical data sharing increases, especially in cross-institutional collaborations and AI model training, these concerns have become more pressing. Protecting patient privacy is both a technical challenge and a legal and ethical obligation. Strict regulations, like the GDPR in Europe and HIPAA in the U.S., have been enacted to address this issue. As a result, developing privacy-preserving methods for medical image analysis has become an urgent research priority \cite{verma2023privacy}.

Over the past decade, numerous privacy-preserving medical image analysis methods have been proposed. These methods not only secure data during sharing and usage but also mitigate the risk of data breaches, fostering greater patient trust. As research in this field expands, several surveys have emerged. For example, Guan et al. \cite{guan2024federated} and Sohan et al. \cite{sohan2023systematic} review federated learning (FL) in medical imaging, illustrating how institutions can collaboratively train AI models while preserving privacy. Liu et al. \cite{liu2024survey} focuses on differential privacy (DP) for safeguarding medical data, while Jin et al. \cite{jin2019review} summarizes blockchain-based solutions for secure, privacy-preserving medical data sharing. Lata et al. \cite{lata2023deep} and Churi et al. \cite{churi2019systematic} analyze various encryption methods, assessing their strengths and weaknesses in medical image encryption. Verma et al. \cite{verma2023privacy} evaluates 12 non-storage, privacy-preserving continuous learning algorithms for medical image classification in quasi-incremental learning scenarios. However, these reviews often focus on specific technologies and overlook their applications to distinct medical image analysis tasks like diagnostic imaging, pathology analysis, and telemedicine, as well as the unique challenges and solutions in these contexts. Although Khalid et al. \cite{khalid2023privacy} and Guerra et al. \cite{guerra2023privacy} cover a wide range of privacy-preserving technologies across healthcare, they lack a detailed comparison and in-depth discussion specific to medical image analysis tasks.

\textbf{Aims and Conributions} The main contribution of this review is to provide a comprehensive perspective on the application of privacy-preserving techniques in medical image analysis. The review is innovatively organized according to the specific challenges in different medical image analysis applications and their corresponding solutions, making the application of technology directly related to practical problems, so that researchers and practitioners can have a clearer understanding of the available options in practical tasks. This addresses the gaps in the current research field. In addition, we also discuss the advantages and limitation of current methods and emerging trends such as zero-knowledge proofs and secure multi-party computation, providing insights for future research. With the continuous development of artificial intelligence and medical image analysis, the need for strong privacy protection will become increasingly important. A systematic review of current methods and prospects for future developments can not only guide future research but also support the practical deployment of privacy-preserving techniques in medical imaging.

\section{Common Techniques}
\vspace{-0.1in}
\subsection{Data Encryption}
Data encryption is a widely used privacy protection method in medical image analysis, ensuring the security of data during transmission and storage by encoding it to prevent unauthorized access or breaches. The two primary types are \textit{static encryption}, which secures stored data, and \textit{in-transit encryption}, which protects data during network transmission.

In medical image analysis, encryption plays a key role in privacy-preserving \cite{hasan2021lightweight}. When hospitals share imaging data, encryption ensures that intercepted data remains unreadable to unauthorized parties. Advanced techniques like symmetric and asymmetric encryption further enhance security and efficiency. \textit{Symmetric encryption} uses the same key for encryption and decryption, providing speed and suitability for large medical image datasets \cite{bisht2023efficient}, though it poses challenges in key management. \textit{Asymmetric encryption} employs a public-private key pair, providing enhanced security but at the cost of higher computational demands, making it less efficient for large datasets \cite{ningthoukhongjam2024medical}. While encryption improves privacy, it involves trade-offs, particularly with significant computational loads for large datasets or resource-constrained environments. Symmetric encryption struggles with key management, and asymmetric encryption, though secure, can impact performance. Hybrid approaches address these issues by using asymmetric encryption for key exchange and symmetric encryption for data. Research continues to optimize encryption methods to balance security and efficiency, with homomorphic encryption being a promising solution, to be discussed next.

\vspace{-0.1in}
\subsection{De-identification and Anonymization}
De-identification and anonymization are commonly used to protect privacy in medical image sharing and processing \cite{chevrier2019use}. \textit{De-identification} involves removing personally identifiable information such as names, ID numbers, and addresses, as well as metadata like birthdates and hospital identifiers from image files. \textit{Anonymization}, a more stringent approach, erases both direct and indirect personal identifiers, making re-identification nearly impossible, even through cross-referencing with other data sources. This is particularly important for large-scale data sharing or public datasets.

In medical image analysis, de-identification preserves data integrity, allowing analysis without significant loss of accuracy. However, it is not foolproof, as re-identification risks remain when anonymized data is cross-referenced with other datasets, especially if the images contain rare or unique features. For instance, specific lesion patterns or distinctive physiological traits might inadvertently reveal a patient's identity. On the other hand, anonymization, while offering stronger privacy protection, may reduce the usability of the data, as removing too much information can affect clinical research or diagnostic accuracy. Furthermore, advances in AI could potentially infer personal information from image patterns, raising concerns about the security of even anonymized data.

\vspace{-0.1in}
\subsection{Differential Privacy}
DP is an emerging technique gaining traction in medical image analysis for its ability to enable statistical analysis and DL model training without compromising individual privacy \cite{ziller2021medical}. It works by adding random noise to mask individual data points while preserving the overall utility of the dataset. The key principle of DP is that analyses on two nearly identical datasets (differing by only one data point) should produce indistinguishable results, ensuring that individual data is concealed while maintaining the dataset’s statistical validity. 

In medical image analysis, DP is primarily applied in two areas: \textit{data sharing} and \textit{DL model training}. DP is valuable for sharing medical image data between institutions, as it adds noise to obscure individual patient features, minimizing the risk of re-identification while allowing secure data exchange. In DL model training, DP prevents models from memorizing sensitive details by adding noise to gradient updates, ensuring that AI models trained on medical images, do not reveal patient-specific characteristics \cite{ziller2021medical}. DP provides strong privacy guarantees, reducing risks of data leakage and re-identification without compromising data integrity, making it particularly suitable for large-scale medical image analysis. It can also be combined with other privacy-preserving techniques like federated learning and homomorphic encryption for enhanced protection. However, DP's introduction of noise can affect data accuracy. While the noise can often be averaged out, excessive noise may reduce precision. Its effectiveness depends on careful noise calibration to balance privacy protection and data utility.

\vspace{-0.1in}
\subsection{Homomorphic Encryption}
Homomorphic encryption allows computations to be performed directly on encrypted data without decryption \cite{wood2020homomorphic}, enabling third parties to process data while preserving both privacy and utility. The key principle is that operations on encrypted data produce the same results as if performed on unencrypted data. There are two types: \textit{Partial Homomorphic Encryption (PHE)}, which allows either addition or multiplication on encrypted data, and \textit{Fully Homomorphic Encryption (FHE)}, which supports both operations in any combination.

In medical image analysis, homomorphic encryption is particularly useful for privacy-preserving AI model training \cite{kumar2022blockchain}. It allows institutions to collaborate on machine learning tasks without exposing sensitive patient data, enabling secure training on encrypted data. The key advantage of homomorphic encryption is its ability to perform secure computations on encrypted data, making it ideal for sensitive medical image analysis. However, FHE is computationally intensive and slow, especially for large datasets, which can hinder the efficiency of model training. PHE offers better performance but supports only a single operation, limiting its use in more complex tasks. These limitations make homomorphic encryption less suited for real-time medical image processing.

\vspace{-0.1in}
\subsection{Federated Learning}
FL is a distributed DL approach that allows multiple institutions to collaboratively train AI models without sharing original data \cite{yin2021comprehensive}. Each institution trains the model locally and exchanges only model parameters (e.g., gradients), ensuring privacy and security, making FL ideal for privacy-sensitive medical image analysis. It also addresses the challenge of data silos, enabling decentralized training across institutions without violating privacy, and improving AI performance by leveraging combined datasets \cite{guan2024federated}.

FL’s key strength is its privacy-preserving nature. By keeping data local and sharing only model updates, it reduces risks associated with centralized data storage and transmission, making it well-suited for sensitive medical data. However, FL faces challenges such as limited network bandwidth and computational resources, which can affect update efficiency and model performance. Integrating heterogeneous data from different institutions is also difficult due to varying equipment, standards, and data formats. Moreover, while FL prevents direct data sharing, there is still a risk of privacy breaches, as attackers may infer sensitive information from model updates.

\vspace{-0.1in}
\subsection{Generative Adversarial Networks (GANs)}
GANs are DL models that generate highly realistic synthetic data through an adversarial process between two networks: the generator and the discriminator. A key application of GANs in privacy-preserving is creating synthetic data to replace real data, thus safeguarding privacy.

In medical image analysis, GANs generate synthetic images, reducing reliance on real data and enhancing privacy-preserving \cite{yi2019generative}. Hospitals and research institutions can train GANs to create synthetic various medical images that mimic the statistical properties of real image without containing sensitive information. This synthetic data is useful for AI model training, algorithm testing, and cross-institutional sharing, minimizing the risk of exposing patient data \cite{chen2022generative}. The key advantage of GANs is their ability to produce high-quality synthetic data, enabling privacy-preserving data sharing and accurate AI model training. However, GANs have limitations: synthetic data may lack critical details, particularly in complex disease-related features, potentially biasing AI models. GAN training is also computationally intensive, and the data generated cannot fully eliminate re-identification risks in certain cases.

\vspace{-0.1in}
\subsection{Others}
In medical image analysis, several image-based privacy-preserving techniques are to obscure patient information while preserving key diagnostic features, including image obfuscation \cite{gaudio2023deepfixcx}, deformation \cite{andrew2020efficient}, and noise addition \cite{huang2022privacy}. 
\textit{Image Obfuscation} disrupts pixel arrangements or applies nonlinear transformations, making identification difficult while preserving statistical properties for AI model training \cite{gaudio2023deepfixcx}.
\textit{Image Deformation} alters the image's geometry through scaling, rotation, or distortion, protecting sensitive information while retaining essential diagnostic features for machine learning \cite{andrew2020efficient}.
\textit{Image Noise Addition} introduces random noise to hide identity-related details. The noise must be carefully calibrated to maintain diagnostic quality while preventing unauthorized reconstruction \cite{huang2022privacy}.

\section{Applications in Medical Image Analysis} 
\begin{figure}
\vspace{-0.2in}
\includegraphics[width=\textwidth]{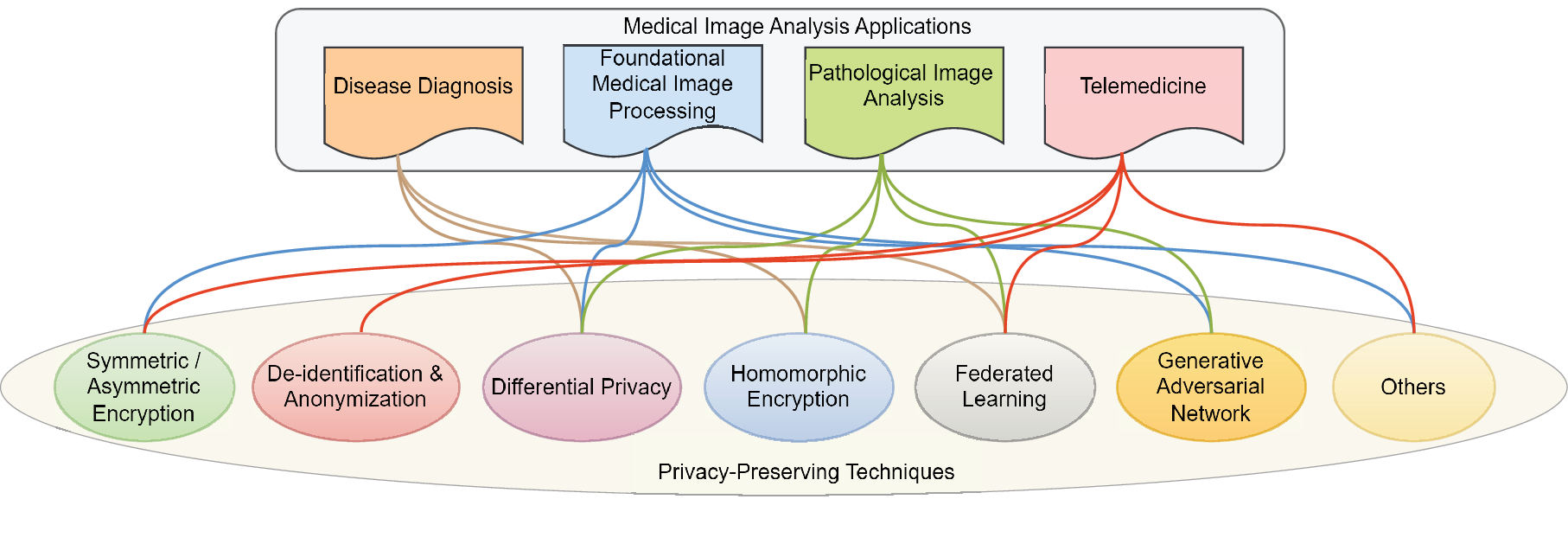}
\vspace{-0.3in}
\caption{Privacy-preserving techniques in different medical image analysis applications.} \label{disgram_overview}
\end{figure}

\vspace{-0.3in}
\subsection{Disease Diagnosis}
Disease diagnosis, such as for cancer and heart disease, heavily depends on analyzing medical images like CT, MRI, and X-rays. These images contain both pathological information and sensitive personal health data, creating significant privacy challenges. The need for large datasets to train AI models increases the risk of privacy leakage when sharing data. Thus, the key issue is how to enable data exchange and model training while protecting patient privacy.

\textbf{Differential privacy} is widely used for privacy-preserving medical imaging diagnosis. By adding noise to datasets, it reduces the risk of re-identification during model training and data sharing. Kaissis et al. \cite{kaissis2021end} developed a framework combining differential privacy with secure aggregation for pediatric chest X-ray classification. Faisal et al. \cite{faisal2022generating} proposed a method to generate high-quality synthetic medical images while preserving privacy. Ziller et al. \cite{ziller2021medical} introduced a framework for differentially private deep learning on pediatric pneumonia and liver diagnosis.

\textbf{Federated learning} is another approach that protects privacy in multi-center disease diagnosis, such as cancer and pneumonia. Zhou et al. \cite{zhou2024personalized} applied FL to classify healthy individuals and patients from lung CT images, while Yang et al. \cite{yang2023study} implemented privacy-preserving FL for chest X-ray classification. Riedel et al. \cite{riedel2023resnetfed} proposed ResNetFed, a federated DL model for privacy-preserving pneumonia detection from COVID-19 chest X-rays, and Lai et al. \cite{lai2024bilateral} introduced the bilateral proxy framework (BPF) for more accurate and reliable diagnosis in federated settings. These studies demonstrate that FL can protect data privacy while improving model performance.

\textbf{Homomorphic encryption} is frequently used in telemedicine diagnosis, because it allows computations to be performed on encrypted data, preventing data leakage during transmission and processing. Kumar et al. \cite{kumar2022blockchain} developed a privacy-preserving model aggregation system for COVID-19 screening using chest X-rays, based on homomorphic encryption. Kolhar et al. \cite{kolhar2023privacy} applied convolutional Bi-LSTM networks to evaluate fully homomorphic encrypted medical images and achieved the detection for five chest diseases. Panzade et al. \cite{panzade2024medblindtuner} proposed MedBlindTuner, which utilized fully homomorphic encryption and a data-efficient image transformer to detect diseases in dermoscopic and 3D CT images, achieving results comparable to non-encrypted scenario.

\vspace{-0.1in}
\subsection{Foundational Medical Image Processing}
Basic medical image processing tasks like segmentation, detection, image registration and compression provide essential data for diagnosis. These tasks involve handling high-resolution images, requiring precise edge detection and segmentation, and are computationally intensive. Privacy-preserving in this area faces unique challenges. For example, in image compression, maintaining privacy without sacrificing performance is critical, while in detection and segmentation, privacy techniques must not reduce accuracy, as this impacts diagnostic outcomes. Common methods used include encryption, GANs, DP, and image obfuscation.

\textbf{Encryption} Liu et al. \cite{liu2019privacy} proposed a lightweight privacy-preserving Faster R-CNN framework for object detection in medical images, which effectively secures Faster R-CNN computations through additive secret sharing and edge computing. Zhu et al. \cite{zhu2024mp} introduced a novel framework for secure medical image segmentation that requires only one encrypted data transmission, encrypting image data into ciphertext, and employs an improved U-Net for segmenting the encrypted images. This method achieved robust and informative segmentation on cardiac MRI and CTPA datasets. Taiello et al. \cite{taiello2024privacy} developed a privacy-preserving image registration technique that operates on encrypted images without exposing the underlying data, optimizing registration through gradient approximation. 

\textbf{GANs} Kim et al. \cite{kim2019privacy} presented a privacy-preserving network based on adversarial learning for MRI segmentation, which obfuscates patient identity while retaining sufficient task-relevant information. Their experiments showed that this approach could severely distort input images while maintaining high segmentation accuracy. Montenegro et al. \cite{montenegro2021privacy} proposed a privacy-preserving GAN for tumor segmentation, which balances privacy and interpretability by generating privacy-preserving images that avoid exposing patient identity while maintaining clarity for diagnostic interpretation. 

\textbf{Differential privacy} Ziller et al. \cite{taiello2024privacy} introduced a differential privacy deep learning framework for liver segmentation in medical image segmentation tasks as part of the Medical Segmentation Decathlon.

\textbf{Image obfuscation} Gaudio et al. \cite{gaudio2023deepfixcx} proposed DeepFixCX, which compresses images by removing or blurring spatial and edge information, improving DL classification performance by 2\% in detecting glaucoma and cervical types.

\vspace{-0.1in}
\subsection{Pathological Image Analysis}
Pathological image analysis is a critical domain in medical image analysis, extensively used for cancer diagnosis, tissue analysis, and lesion detection. Compared to other medical images, pathological images possess extremely high resolution and intricate details, making them particularly susceptible to re-identification risks. Privacy-preserving techniques in this field must ensure that the integrity of these fine details is not compromised. Additionally, the challenges of data acquisition and annotation in pathology highlight the need for optimizing data usage while maintaining privacy protection. Common privacy-preserving methods in this field include FL, GANs, DP, and homomorphic encryption.

\textbf{Federated learning} Gkillas et al. \cite{gkillas2024privacy} employed FL for skin cancer diagnosis from images. Their results demonstrated that this approach achieves the same accuracy as traditional DL models, while significantly enhancing privacy, supporting heterogeneous devices, and reducing communication requirements. Lu et al. \cite{lu2022federated} applied privacy-preserving FL to gigapixel whole-slide images in computational pathology, using weakly supervised attention-based multi-instance learning and differential privacy. Their study showed that FL can effectively develop accurate weakly supervised models from distributed data without direct data sharing. Additionally, differential privacy was ensured through the generation of random noise, further enhancing privacy protection.

\textbf{GANs} Shen et al. \cite{shen2022federated} introduced a conditional GAN within a FL paradigm to enhance data privacy and security. Their experiments indicated that the model's performance closely matched traditional centralized learning approaches, while ensuring stronger privacy protection. Xiong et al. \cite{xiong2024privacy} developed a novel dataset distillation method using a multidimensional matching approach to generate synthetic datasets that retain the utility of the original medical datasets while enhancing privacy. Their method, tested on the colon pathology benchmark dataset, not only improved model performance but also significantly enhanced the privacy of the distilled images.

\textbf{Differential privacy} Kumar et al. \cite{kumar2021novel} proposed a secure and efficient framework for cancer diagnosis using histopathology image classification, combining DP with secure multi-party computation. Their approach was validated on datasets comprising histopathology images of canine mammary tumors and human breast cancer, demonstrating its effectiveness in preserving privacy while maintaining diagnostic accuracy.

\textbf{Homomorphic encryption} Al et al. \cite{al2024private} explored the use of FHE in conjunction with DL for private pathology assessment. Their evaluation across various datasets confirmed the system's practicality and effectiveness in protecting sensitive medical data.

\vspace{-0.1in}
\subsection{Telemedicine}
With the rapid advancement of telemedicine and cloud computing, medical image sharing is essential for cross-institutional collaboration, real-time diagnosis, and data analysis. However, these processes heighten the risks of data leakage and unauthorized access. Key privacy challenges include the absence of unified privacy standards for cross-institutional data sharing, increasing the potential for breaches, and the security vulnerabilities associated with cloud-based image analysis. To address these challenges, effective privacy-preserving techniques include image deformation, obfuscation and noise addition, encryption, and FL.

\textbf{Image deformation/obfuscation/Noise addition} Andrew et al. \cite{andrew2020efficient} proposed a secure DL approach called deformation learning to safeguard image privacy in medical image analysis. Their method allows data providers to distort images using a deformation component, removing privacy-sensitive information. The resulting image, unrecognizable to humans, is then sent to the service provider, where deep learning algorithms, enhanced by convolutional layers, process the deformed data without performance loss. Similarly, Kim et al. \cite{kim2020privacy} introduced a method to protect patient identity by applying pseudo-random nonlinear deformation to input images, creating a proxy image for server-side processing. Sun et al. \cite{sun2024plaintext} developed the Frequency Domain Exchange Style Fusion method, which fuses plaintext medical image features with a host medical image. These alternative images are used during model training, ensuring anonymous analysis without plaintext data during the training and inference stages. Huang et al. \cite{huang2022privacy} suggested enhancing privacy protection for medical records by adding noise and then applying denoising techniques to the same records.

\textbf{Encryption} Xiang et al. \cite{xiang2023bmif} introduced the Blockchain Medical Image Fusion framework, which utilizes homomorphic encryption to securely train fusion models on encrypted medical data. Li et al. \cite{li2024pmir} proposed the Privacy-preserving Medical Image Retrieval scheme, in which encrypted image data and indexes are uploaded to a cloud server. Rigorous security analysis confirmed that PMIR effectively resists various known security threats. Moreover, experimental results demonstrated that PMIR reduces communication overhead, providing users with a seamless and secure search experience.

\textbf{Federated learning} Yu et al. \cite{yu2023pomic} proposed a flexible outsourcing scheme for privacy-preserving medical image classification on the cloud using convolutional neural networks. Their approach reduces computational and storage burdens on the user side while ensuring security and efficiency, as demonstrated by experimental implementation.

%\vspace{-0.1in}
\section{Advantages, Limitations, and Future Directions}
Privacy-preserving techniques are crucial in medical image analysis, providing effective tools to safeguard patient data. Below is a summary of their key advantages, limitations, and future directions.

\textbf{Advantages}
Technologies like FL enable collaborative learning across institutions without sharing data, protecting privacy while maintaining model performance. GANs generate synthetic data for AI training, reducing the risk of exposing sensitive information. Homomorphic encryption allows computations on encrypted data, ensuring confidentiality during processing. DP enhances security by adding noise, preventing re-identification risks during data analysis.

\textbf{Limitations}
Privacy technologies, such as differential privacy, may reduce model performance by introducing noise, particularly in tasks requiring high-resolution images or precise segmentation. Methods like homomorphic encryption, while highly secure, incur significant computational complexity, especially in large-scale cloud-based or telemedicine applications, where processing times and costs can be prohibitive. Although federated learning and differential privacy offer theoretical scalability, in practice, issues like high communication overhead and data synchronization challenges hinder its scalability. Moreover, these methods often require substantial computing and storage resources, limiting their use in resource-constrained environments.

\textbf{Development Trends} (1) Future privacy technologies will prioritize personalization, adapting solutions to specific medical scenarios like telemedicine and cross-institutional data sharing, especially as AI in disease diagnosis advances. (2) A critical focus will also be on balancing security and usability, developing lightweight, efficient solutions that ensure privacy protection in resource-limited environments. Reducing the complexity of homomorphic encryption and improving real-time data processing will be essential to achieving this balance.

\textbf{Future Directions} (1) Enhancing differential privacy: while differential privacy protects data, the noise introduced can impair model accuracy. Future research will explore dynamic noise adjustment tailored to different medical tasks, aiming to balance high accuracy with robust privacy. (2) Optimizing federated learning: despite its success in multi-institutional data sharing, federated learning still faces challenges related to communication overhead and model synchronization. Research will focus on developing efficient communication protocols and model update strategies to reduce the cost and complexity of multi-party collaboration. (3) Zero-knowledge proofs: this technique, which verifies data authenticity without revealing its content, has strong potential for privacy protection. Future research may integrate zero-knowledge proofs with existing technologies to enhance privacy and data verification in medical image analysis. (4) Multi-Party computation: multi-party computation enables collaborative tasks without sharing data. Further research will explore its application in medical image analysis, particularly in scenarios requiring secure, cross-institutional collaboration.

\section{Conclusion}
This review explored key privacy-preserving methods in medical image analysis. While they are essential for privacy-preserving, challenges such as performance degradation, computational complexity, and scalability persist. Future research should focus on creating adaptable privacy frameworks that balance security with usability, lower computational costs, and enhance real-time processing. Emerging technologies like zero-knowledge proofs and multi-party computation present promising solutions for improving privacy in medical imaging while maintaining efficiency and accuracy.

\bibliographystyle{splncs04}
\bibliography{mybib}

\end{document}